\pdfoutput=1

\documentclass[11pt]{article}

\usepackage{EACL2023}

\usepackage{times}
\usepackage{latexsym}

\usepackage[T1]{fontenc}

\usepackage[utf8]{inputenc}

\usepackage{microtype}

\usepackage{makecell}
\usepackage{multirow}
\usepackage[normalem]{ulem}
\usepackage{amssymb}
\usepackage{graphicx}

\usepackage{amsmath}
\mathchardef\mhyphen="2D 
\usepackage{microtype}

\usepackage{enumitem} 
\usepackage{xcolor}
\usepackage{colortbl}

\usepackage{soul}

\newcommand{\smm}[1]{{\color{black} #1}} 
\newcommand{\bl}[1]{{\color{black} #1}} 
\newcommand{\rd}[1]{{\color{black} #1}} 
\newcommand{\gr}[1]{{\color{black} #1}}
\newcommand{\fl}[1]{{\color{black} #1}}

\usepackage{inconsolata}

\title{What Makes Sentences Semantically Related?\\ 
 A Textual Relatedness Dataset and Empirical Study}

\author{\hspace*{-16mm} Mohamed Abdalla \\
	  \hspace*{-16mm}   Institute for Better Health \\ \hspace*{-16mm} \& University of Toronto\\
	 \hspace*{-16mm}     {\tt msa@cs.toronto.edu}  \\
   \And 
   \hspace*{-16mm} Krishnapriya Vishnubhotla \\
	\hspace*{-16mm}     University of Toronto\\
	 \hspace*{-16mm}     {\tt vkpriya@cs.toronto.edu}
   \And 
        \hspace*{-4mm} Saif M. Mohammad\\
	    \hspace*{-4mm} National Research Council Canada\\
	    \hspace*{-4mm} {\tt saif.mohammad@nrc-cnrc.gc.ca}  \\}

\begin{document}
\maketitle

\begin{abstract}
The degree of semantic relatedness of two units of language has long been considered fundamental to understanding meaning. Additionally, automatically determining relatedness has many applications such as question answering and summarization.  However, prior NLP work has largely focused on semantic similarity, a subset of relatedness, because of a lack of relatedness datasets.
In this paper, we introduce a dataset for Semantic Textual Relatedness, \textit{STR-2022}, that has 5,500 English sentence pairs manually annotated using a comparative annotation framework, resulting in fine-grained scores. We show that human intuition regarding relatedness of sentence pairs is highly reliable, with a repeat annotation correlation of 0.84. We use the dataset to explore questions on what makes sentences semantically related. We also show the utility of STR-2022 for evaluating automatic methods of  sentence representation and for various downstream NLP tasks.

Our dataset, data statement, and annotation questionnaire can be found at: \url{https://doi.org/10.5281/zenodo.7599667}.
\end{abstract}

\section{Introduction}
\label{sec:intro}
 \setitemize[0]{leftmargin=*}
 \setenumerate[0]{leftmargin=*}

The semantic relatedness of two units of language
is the degree to which they are close in terms of their meaning 
\cite{mohammad2008measuring,mohammad2012distributional}. 
The linguistic units can be words, phrases, sentences, etc.
Though our 
intuition
of semantic relatedness is dependent on many factors such as the context of assessment, age, and socio-economic status \citep{harispe2015semantic}, it is argued that a consensus can usually be reached for many pairs \citep{harispe2015semantic}.  Consider the two sentence pairs in Table \ref{tab:distance-ex}.  Most speakers of English will agree that the sentences in the first pair are closer in meaning to one another than \gr{those} in the second. When judging the semantic relatedness between two sentences, humans generally look for commonalities in meaning: whether they are on the same topic, express the same view,  originate from the same time period, 
one elaborates on (or follows from) the other, etc.


\begin{table}[t]
	\centering
	\small
	\begin{tabular}{l}
		\textbf{Pair 1:} a. \textit{There was a lemon tree next to the house.} \\
		$\;\;\;\;\;\;\;\;\;\;\,$ b. \textit{The boy enjoyed reading under the lemon tree.} \\[3pt]
		\textbf{Pair 2:} a. \textit{There was a lemon tree next to the house.} \\
		$\;\;\;\;\;\;\;\;\;\;\,$ b. \textit{The boy was an excellent football player.} 
	\end{tabular}
	\vspace*{-2mm}
	\caption{\label{tab:distance-ex} 
	Most people will agree that the sentences in pair 1 are more related than the sentences in pair 2.}
	\vspace*{-4mm}
\end{table}

The 
semantic relatedness of two units of language has long been considered fundamental to understanding meaning \cite{halliday1976cohesion,miller1991contextual}; 
given how difficult it has been to define meaning, a natural approach to 
\gr{get at the} meaning \gr{of a unit is to determine how close it is to other units.}
\smm{Thus, unsurprisingly,} automatically determining 
relatedness has many applications such as question answering, text generation, and summarization \bl{(more discussion in \S\ref{sec:utility})}.  

However, prior NLP work has focused on semantic similarity (a small subset of semantic relatedness), largely because of a dearth of datasets
on relatedness. 
The few relatedness datasets that exist are only for word pairs \cite{rubenstein1965contextual,radinsky2011word} or phrase pairs \cite{asaadi2019big}. Further, most existing datasets were annotated, one item at a time, using coarse rating labels such as integer values between 1 and 5\@ representing coarse degrees of closeness. It is well documented that such approaches suffer from inter- and intra-annotator inconsistency, scale region bias, and issues arising due to the fixed granularity \cite{presser2004questions}. 
Further, the notions of \textit{related} and \textit{unrelated} have fuzzy boundaries. Different people may have different \gr{intuitions} of where such a boundary exists. 
\rd{Finally, for some tasks, it is more appropriate to train on a dataset of relatedness than similarity. (\S\ref{sec:rel-sim} discusses how relatedness and similarity are different.)}

In this paper, we present the first manually annotated dataset of sentence--sentence semantic relatedness. It includes fine-grained scores of relatedness from 0 (least related) to 1 (most related) for 5,500 English sentence pairs. The sentences are taken from diverse sources and thus also have diverse sentence structures, varying amounts of lexical overlap, and varying formality.

The relatedness scores were obtained using a \textit{comparative} annotation schema: 
two (or more) items are presented together and the annotator has to determine which is greater with respect to the metric of interest. Since annotators are making relative judgments, the limitations discussed earlier for rating scales are greatly mitigated. Importantly, such annotations do not rely on arbitrary boundaries between arbitrary categories such as ``\gr{strongly} related'' and ``\gr{somewhat} related''.  

\noindent We use the relatedness dataset to explore:
\vspace*{-3mm}
\begin{enumerate}
    \item To what extent do speakers of English \rd{intuitively} agree on the relatedness of 
    sentences? (\S \ref{sec:relatedness_reliability})
    \vspace*{-2mm}
    \item What makes two sentences more related? 
    (\S \ref{sec:pos_analysis})
\vspace*{-2mm}
    \item How well do existing approaches of sentence representation
    capture 
    semantic relatedness 
    (by placing related sentence pairs closer to each other in vector space)? 
    (\S \ref{sec:sentenceRep})
\vspace*{-2mm}
    \item \bl{How can an improved annotation schema to capture relatedness benefit other NLP tasks? (\S \ref{sec:utility})}
\end{enumerate}
\vspace*{-2mm}
\noindent We refer to our dataset as \textit{STR-2022}, and the task of 
predicting relatedness between sentences as the \textit{Semantic Textual Relatedness (STR)} task. Data, data statement,   
and annotation questionnaire 
are  made 
available\footnote{\url{doi.org/10.5281/zenodo.7599667} or \\ \url{https://github.com/Priya22/semantic-textual-relatedness} or \\ \url{https://huggingface.co/datasets/vkpriya/str-2022}}.


\section{\rd{Related Work and Our Approach to Annotating for Semantic Relatedness}}

The three subsections below discuss key ideas 
\smm{from past work on annotating relatedness and similarity,}
existing datasets, and comparative annotation, respectively.
\smm{Notably, each of these subsections also discusses 
how relevant past work has influenced our approach to data annotation.}

\subsection{Annotating Relatedness and Similarity}
\label{sec:rel-sim}

\bl{Semantic relatedness and semantic similarity are two ways to \smm{explore} closeness of meaning.} Two terms are considered semantically similar if there is a synonymy, hyponymy, 
or troponymy relation between them (examples include \textit{doctor--physician} and \textit{mammal--elephant}). Two terms are considered to be semantically related if there is any lexical semantic relation at all between them. Thus, all similar pairs are also related, but not all related pairs are similar. For example, \textit{surgeon--scalpel} and \textit{tree--shade} are related, but not similar.

Analogous to term pairs, two sentences are considered semantically similar when they have a paraphrasal or entailment relation. Determining such an equivalence of meaning is useful in NLP tasks such as text summarization 
and plagiarism detection.
Semantic Relatedness, however, accounts for all of the commonalities that can exist between two sentences \cite{halliday1976cohesion,morris1991lexical}. For example, the sentences in Table \ref{tab:distance-ex} Pair 1 are highly related, but they are not paraphrases or entailing. This expands the scope of the measure 
to include aspects such as the relatedness between their topics, their styles, etc. 

However, 
\smm{because semantic relatedness involves innumerable classical and ad-hoc semantic relationships, it is markedly more complex than
semantic similarity, and}
there are no widely agreed upon linguistic theories \smm{or guidelines for judging} relatedness. This presents a challenge for gathering annotations; 
one can either: 
(i) construct their own codified instructions on how to judge semantic relatedness under various scenarios
(e.g., overlapping sentence structure, relatedness of topic, etc.),
\smm{at the risk of artificially over-simplifying the task}
or
(ii) abstain from explicitly and comprehensively defining relatedness 
\smm{for numerous types of sentence pairs}, relying instead on 
\smm{a simple description of relatedness, a few examples, and framing the task in relative terms.}\footnote{\smm{Recall that for Table 1, we were able to judge relative relatedness without explicit instruction on how to judge relatedness.}} 
In this work, we chose 
the latter. This allows us to: 
(i) 
\smm{determine} the extent to which human intuition of relatedness 
is reliable 
and 
(ii) use the resulting 
dataset to empirically determine 
\smm{what makes sentences semantically related}.

\subsection{Existing Relatedness and Similarity Data}

\rd{\bl{Existing} datasets created for sentence pair similarity (e.g., STS \cite{agirre2012semeval,agirre2013sem,agirre2014semeval,agirre2015semeval,agirre2016semeval}, MRPC \cite{dolan2005automatically}, and LiSent \cite{li2006sentence}) 
ask annotators to choose among coarse similarity labels. 
This 
\smm{leads to}
information loss 
\smm{and} makes annotation difficult because distinctions between categories are often not clear; for example, the STS 2012--2016 questionnaires ask annotators to make the distinction between \textit{2: not equivalent but share some details} and \textit{1: not equivalent, but are on the same topic}, which is often not straightforward. 
\smm{Further,} despite claiming to determine semantic similarity, the descriptions of categories 1 and 2 incorporate aspects of semantic relatedness --- an amalgamation muddying the waters with respect to the phenomenon being annotated. 
Such an amalgamation 
\smm{is also seen in} the SICK \cite{marelli2014sick} dataset which combines a labeling scheme from STS with 
those
about entailment and contradiction. 
\smm{These datasets have helped make progress in the field, but there is a need for relatedness datasets
obtained strictly from relatedness judgments as opposed to a hybrid involving artificially created categories for similarity and entailment.}
}
For our annotations, \smm{we avoid fuzzy ill-defined categories}, and 
rely instead on the intuitions of fluent English speakers to judge \textbf{relative rankings} of sentence pairs by relatedness.

\subsection{Comparative Annotations}
\label{sec:comparative_annotations}

The simplest form of comparative annotations is paired comparisons \cite{thurstone1927law,david1963method}. 
Annotators are presented with pairs of examples and 
asked to choose which item is greater 
with respect to the property of interest 
(relatedness, sentiment, etc.). 
The choices 
\smm{are then}
used to generate an ordinal ranking of items. 
\smm{Paired comparison avoids a number of biases, but it}
requires a large number of annotations 
($N^2$, where $N=$ \# items). 

Best--Worst Scaling (BWS) is a comparative annotation schema that builds on pairwise comparisons and 
requires fewer labels \cite{louviere1991best}. 
Annotators are 
given
$n$ items at a time (for our work, $n$ = 4 and an \textit{item} is a pair of sentences). They are instructed to choose the best (i.e., most related) and worst (i.e., least related) item. Annotation for each 4-tuple provides us with five pairwise inequalities. For example if $a$ is marked as most related and $d$ as least related, then we know that $a>b, a>c, a>d, b>d$, and $c>d$. 
\smm{These inequalities can be used to} calculate real-valued scores, and thus an ordinal ranking of items, using a simple counting mechanism \cite{orme2009maxdiff,flynn2014best}: 
the fraction of times an item was chosen as the best (most related) minus the fraction of times the item was chosen as the worst (least related). Given $N$ items, reliable scores are obtainable from about $2N$ 4-tuples \cite{bws-naacl2016,kiritchenko2017best}. 

\section{Creating STR-2022}

Dataset creation included several steps: curating sentence pairs for annotation, designing the questionnaire, crowdsourcing annotations, and aggregating the annotations to obtain relatedness scores.

\subsection{Data Sources}
\label{sec:data_sources}

\rd{Like 
previous work on semantic similarity,} we chose to construct our dataset by sampling sentences from many sources to capture a wide variety of text in terms of sentence structure, formality,
and grammaticality. Pairs of sentences were created from the sampled sentences in a number of ways as described below. The sources are: 
\vspace*{-2mm}
\begin{enumerate}
    \item \textbf{Formality} \cite{rao2018dear}: Pairs of sentences having the same meaning but differing in formality (one formal, one informal).
    \vspace*{-2mm}
    \item  \textbf{Goodreads} \cite{DBLP:conf/recsys/WanM18}: Book reviews from the Goodreads website. 
    \vspace*{-2mm}
    \item \textbf{ParaNMT} \cite{wieting-gimpel-2018-paranmt}: Paraphrases from a machine translation system.
    \vspace*{-2mm}
    \item     \textbf{SNLI} \cite{snli}:  Pairs of premises and hypotheses, created from image captions, for natural language inference. 
    \vspace*{-2mm}
    \item \textbf{STS} \cite{cer2017semeval}: Pairs of sentences with semantic similarity scores. (Integer label responses, 0 to 5, from multiple annotators were averaged to obtain the 
    similarity scores.)
    \vspace*{-2mm}
    \item \textbf{Stance} \cite{mohammad2016dataset}: Tweets labelled for both sentiment (\textit{positive}, \textit{negative}, \textit{neutral}) and stance (\textit{for}, \textit{against}, \textit{neither})
    towards targets (e.g., 
    \textit{Donald Trump}, \textit{Feminism}).
    \vspace*{-7mm}
    \item \textbf{Wikipedia}  Text Simplification Dataset \cite{horn2014learning}: 
    Pairs of Wikipedia sentences and their simplified forms.
\end{enumerate}
\vspace*{-2mm}
\noindent 
From each source, we  sampled sentences that were between 5 and 25 words long. 
\rd{We selected sentence pairs with varying amounts of lexical overlap because randomly sampling sentence pairings would result in mostly unrelated sentences.}
This also allowed us to systematically study the impact of lexical overlap on semantic relatedness. For the paraphrase datasets (Formality, ParaNMT, and Wikipedia), we obtained sentence pairs in two ways: by directly taking the paraphrase pairs (indicated by the suffix {\it \_pp}), and by randomly pairing sentences from two different paraphrase pairs (suffixed by {\it \_r}). The paraphrase pairs were selected at random from the source dataset, whereas the lexical overlap strategy was applied in the creation of the random pairs.
\gr{From STS, 
we randomly sampled 50 sentence pairs
having similarity scores in  $[0\mhyphen 1)$, 50 pairs having scores in $[1\mhyphen 2)$, and so on.
}

\fl{Initially, we sought to annotate 1000 sentence pairs from each source. As our goal was to cover as large a range of relatedness, sentence structures, lengths and topics as possible, we lowered the amount of sentence pairs to obtain the desired variety. For example, SNLI pairs had little variation in sentence length and so we reduced the number of sampled instances; their semantic relatedness also tends to skew towards the higher range, and we aimed to balance this out with a larger number of random, non-paraphrasal pairs from other sources (Formality).
}
Table \ref{tab:data_sources} summarizes key details of the 
sentence pairs in STR-2022. \gr{Further details about the source data and sampling are 
in Appendix \ref{ref:appendix_data}.}

\begin{table}[tbp]
\small
\centering
\begin{tabular}{llr}
\hline
\textbf{Types of Pairs}       &\textbf{Key Attributes} & \textbf{\# pairs} \\ \hline \\[-9pt]

1. Formality                   & paraphrases, style            & \\
$\;\;\;\;$ Formality\_pp           & paraphrases, differ in style  & 300 \\ 
$\;\;\;\;$ Formality\_r            & random pairs                  & 700\\ 
2. Goodreads                & reviews, informal             & 1000 \\
3. ParaNMT                  & automatic paraphrases         & \\ 
$\;\;\;\;$ ParaNMT\_pp          & automatic paraphrases         & 450 \\ 
$\;\;\;\;$ ParaNMT\_r           & random pairs                  & 300  \\ 
4. SNLI                     & captions of images            & 750 \\
5. STS                      & have similarity scores        & 250 \\ 
6. Stance                   & tweet pairs with same ha- &\\ 
                            & shtag, less grammatical              & 750 \\
7. Wikipedia                & formal                        &\\       
$\;\;\;\;\;$ Wiki\_pp       & paraphrases, formal           & 500\\ 
$\;\;\;\;\;$ Wiki\_r        & random pairs, formal          &500\\ \hline 
ALL                         &                               & 5500\\ \hline
\end{tabular}
\caption{\label{tab:data_sources} Summary of sentence pair types in STR-2022.}
\vspace*{-3mm}
\end{table}

\subsection{Annotating For Semantic Relatedness}
From the list of 5,500 sentence pairs, we generated 11,000 unique 4-tuples (each 4-tuple consists of 4 distinct sentence pairs) such that each sentence pair occurs in around eight 4-tuples.\footnote{The tuples were generated using the BWS scripts provided by \citet{kiritchenko2017best}: \url{http://saifmohammad.com/WebPages/BestWorst.html}.}


In our framing of the task, we did not use detailed or technical definitions; rather, we provided brief and easy-to-follow instructions, gave examples, and encouraged annotators to rely on their intuitions of the English language to judge relative closeness in meaning of sentence pairs (similar to Asadi et al.'s (\citeyear{asaadi2019big}) work on bigrams). 
Annotators were asked to judge the ``\textit{closeness in meaning of sentence pairs}''. 
Inspired by early work in linguistics on cohesion in text \cite{halliday1976cohesion}, we also specified that:
``\textit{Often sentence pairs that are more specific in what they share tend to be more related than sentence pairs that are only loosely about the same topic}"
and
"\textit{If a sentence has more than one interpretation, consider that meaning which is closest to the meaning of the other sentence in the pair.}" This is inline with application scenarios where often relatedness is to be determined between sentences from the same document. 
The full 
questionnaire 
is 
included in the supplementary material.

\subsubsection{Crowdsourcing Annotations}
We used Amazon Mechanical Turk (MTurk) for obtaining annotations.\footnote{This project was approved by the first author's Institutional Research Ethics Board (Protocol \#: Masked for review).} 
Each 4-tuple (also referred to as a question) in our MTurk task consists of four sentence pairs. Annotators are asked to choose the (a) most-related, and (b) least-related sentence pairs from among these four options. Each question is annotated by two MTurk workers.\footnote{Pilot studies showed that this results in reliable scores.}

For quality control, the task was open only to fluent speakers of English and those MTurk workers with an approval rate higher than 98\%. Further, we inserted ``Gold Standard'' questions at regular intervals in the task. These questions were manually annotated by all the authors, and had high agreement scores. If an annotator gets a gold question wrong, they are immediately notified and shown the correct answer. This has several benefits, including: keeping the annotator alert and clearing any misunderstandings about the task. Those who scored less than $\sim$70\% on the gold questions were stopped from answering further questions and were paid for their work. All their responses were discarded.

\subsubsection{Annotation Aggregation}

\begin{figure}[tbp]
\centering
\includegraphics[width=0.5\textwidth]{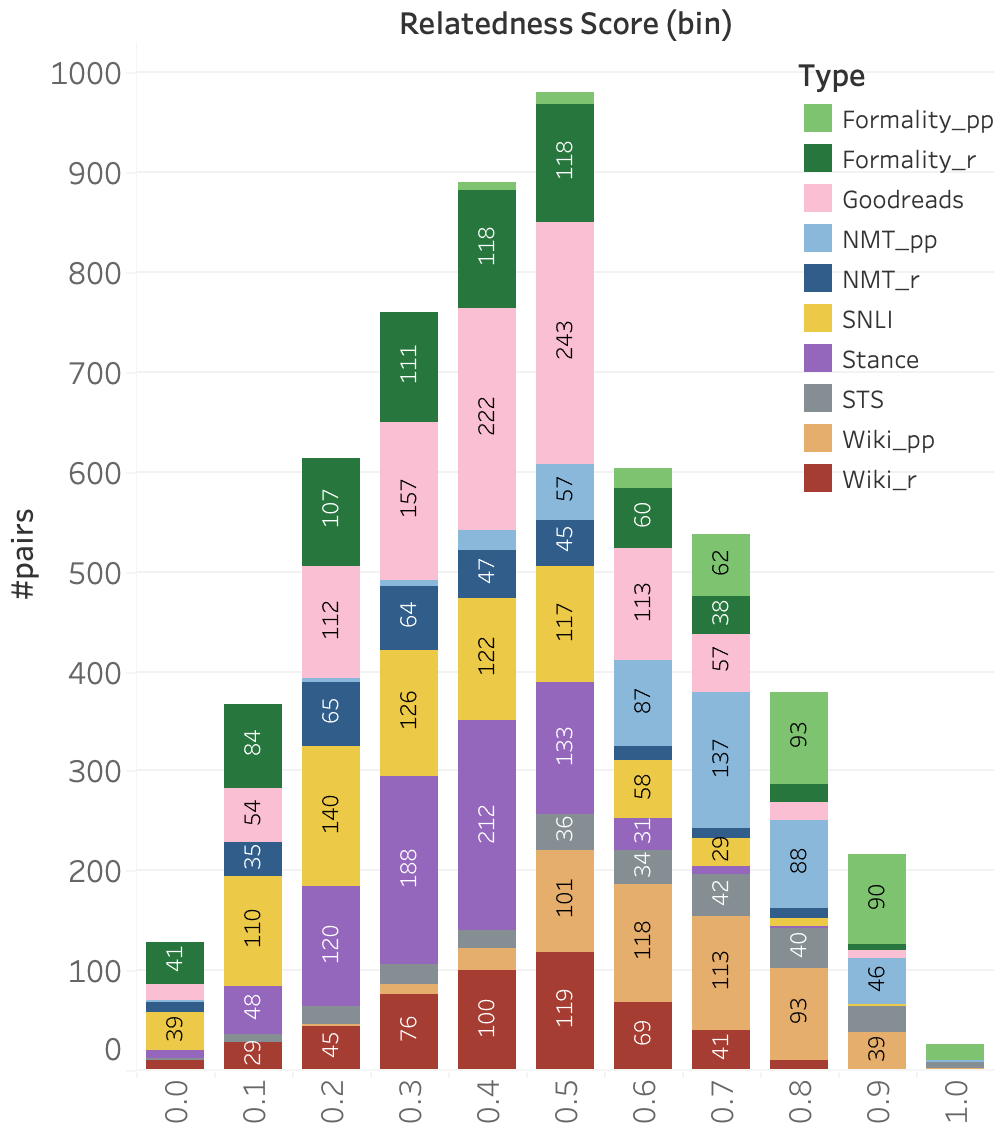}
\vspace*{-7mm}
\caption{\label{img:data_distribution} Histogram of STR-2022 relatedness scores.}
\vspace*{-3mm}
\end{figure}

We aggregate information from various responses by using the counting procedure  discussed in \S \ref{sec:comparative_annotations}. Since relatedness is a unipolar scale, the resulting relatedness score was linearly transformed to fit within a 0--1 scale of increasing relatedness. \fl{Appendix Table \ref{tab:sample-pairs} presents sample sentence pairs from each data source.}

Figure \ref{img:data_distribution} presents a histogram of relatedness scores for STR-2022. Observe that each of the subsets covers a wide range of relatedness scores; 
that the lexical overlap sampling strategy has resulted in a wide spread of relatedness scores;
and 
that 
supposed paraphrases are spread across much of the right half of the relatedness scale. 

\begin{table*}[t]
\centering
{\small
\begin{tabular}{cccccc}
\hline 
\textbf{\# Sentence Pairs} & \textbf{\# Tuples} & \textbf{\# Annotations Per Tuple} & \textbf{\# Annotations} & \textbf{\# Annotators} & \textbf{SHR} \\ \hline\\[-9pt]
5,500                      & 11,000             & 8                                 & 21,936            & 389                   & 0.84         \\ \hline
\end{tabular}
\caption{\label{tab:shr} Annotation statistics \smm{of STR-2022}. SHR = split-half reliability (as measured by Spearman correlation).}
}
\vspace*{-3mm}
\end{table*}

\section{Reliability of Annotations}
\label{sec:relatedness_reliability}

For annotations producing real-valued scores, a commonly used measure of quality and reliability is \textit{split-half reliability} (SHR) \cite{cronbach1951coefficient,kuder1937theory}.  SHR is a measure of the degree to which repeating the annotations would result in similar relative rankings of the items. To measure SHR, annotations for each 4-tuple are split into two bins. The annotations for each bin are used to produce two different independent relatedness scores. Next, the Spearman correlation between the two sets of scores is calculated---a measure of the closeness of the two rankings. If the annotations are reliable then there should be a high correlation. This process is repeated 1000 times and the correlation scores are averaged. 

\smm{
As shown in Table \ref{tab:shr}, STR-2022 has an SHR of 0.84---signifying high annotation reliability. 
This is a key result of this paper. Recall that our annotation guidelines did not hard code the various scenarios of sentence pair types and how they should be judged, but rather were designed to elicit how native speakers of English naturally judge relatedness. The high reliability of annotations, despite this, shows that speakers of a language are inherently consistent in their judgments of relatedness. It also validates our approach as a way to produce high-quality relatedness datasets; which, in turn, can be used to study the mechanisms underpinning relatedness (as we explore in the next Section).
}

 
\rd{
\subsection{STR vs STS}}
\gr{We also conducted  experiments to assess fine-grained rankings of common sentence pairs as per our relatedness scores and as per STS's similarity scores.}
\gr{For each of the sets of 50 sentence pairs taken from STS (with scores in (0--1], (1--2], etc.), }
we calculated the Spearman correlation between the rankings by similarity and rankings by relatedness.
We found that the correlations are  \gr{only 0.25 (weak) and 0.19 (very weak) for} the bins of (1,2] and (3,4], respectively, and only \gr{about 0.49 (moderate)} for the bins of (2,3] and (4,5]. 
Overall, this shows that the fine-grained ranking of items in the STS dataset by similarity differ considerably from that of the 
STR dataset. 

 \setitemize[0]{leftmargin=20pt}

\section{What Makes Sentences More Semantically Related?}
\label{sec:pos_analysis}
The availability of a dataset with human notions of semantic relatedness allows one to explore fundamental 
\gr{aspects of} meaning: for example, what makes two sentences more related? In this section, we 
examine some basic questions. On average, to what extent is \gr{the} semantic relatedness 
\gr{of a sentence pair impacted by presence of}: 
\vspace*{-1mm}
\begin{itemize}
    \item identical words (lexical overlap)? (Q1) 
\vspace*{-3mm}
    \item related words? (Q2)
\vspace*{-3mm}
    \item related words of the same part of speech? (Q3)
\vspace*{-8mm}
    \item related subjects, related objects? (Q4)
\end{itemize}

\subsection{Method}
To explore the questions above, we first
computed relevant measures for Q1 through Q4 (lexical overlap, term relatedness, etc.)
 for each sentence pair in our dataset.
We then calculated the correlations of these scores with the gold relatedness scores.\\[-10pt]
\setitemize[0]{leftmargin=*}

\noindent \textbf{Lexical Overlap}. A simple measure of lexical overlap between two sentences \gr{X} and \gr{Y} is the Dice Coefficient (the number of unique unigrams occurring in both sentences, adjusted by their lengths):
\vspace*{-2mm}
    \begin{equation}
    \frac{2 \times |\,unigram(X) \cap unigram(Y)\,|}{|\,unigram(X)\,|\,+|\,unigram(Y)\,|\,}\\
    \label{eq:dice}
\end{equation}

\noindent \textbf{Related Words:}  
We averaged the embeddings for all the tokens in a sentence and computed the cosine 
between the averaged embeddings for the two sentences in a pair. This roughly captures the relatedness between the terms across the two sentences.\footnote{Other ways to estimate relatedness between sets of words across two sentences may also be used.} Token embeddings were taken from Google's publicly released Word2Vec embeddings trained on the Google News corpus \cite{mikolov2013efficient}.


\noindent \textbf{Related Words with same POS:} The same procedure was followed as for Q2, except that only the tokens for one part of speech \gr{(POS)} at a time were considered. We determined the part-of-speech of the tokens using spaCy \cite{spacy}.\footnote{We used the simple (coarse-grained) UPOS part-of-speech tags: {https://universaldependencies.org/docs/u/pos/}}  \\[-10pt]

\noindent \textbf{Related Subjects and Related Objects:} For Q4, which 
examines the importance of different parts of a sentence,
we employ the same process as Q2, except that for a given sentence: only tokens marked as subject are averaged; and only tokens marked as object are averaged. We use the packages spaCy \cite{spacy} and Subject Verb Object Extractor \cite{svo} to determine all tokens that are  the subject and object.\\[-9pt]

\begin{table}[tbp]
\centering
\small
\begin{tabular}{lrr}
\hline
 \textbf{Question} &\textbf{Spearman} & \textbf{\# pairs} \\ \hline\\[-9pt]
Q1. Lexical overlap       &    0.57       & 5500\\ [3pt]
Q2. Related words - All       & 0.61          & 5500\\ [3pt]
\multicolumn{2}{l}{Q3a. Related words - per POS} \\
$\;\;\;\;\;\;\;\;\;$ PROPN                                                                      & 0.50                                                                                      & 1907                         \\
$\;\;\;\;\;\;\;\;\;$ NOUN                                                                       & 0.45                                                                                     & 4746                         \\
$\;\;\;\;\;\;\;\;\;$ ADJ                                                                        & 0.36                                                                                     & 2236                         \\
$\;\;\;\;\;\;\;\;\;$ VERB                                                                       & 0.31                                                                                     & 3946                         \\
$\;\;\;\;\;\;\;\;\;$ PRON                                                                       & 0.30                                                                                      & 1800                         \\
$\;\;\;\;\;\;\;\;\;$ ADV                                                                        & 0.28                                                                                     & 1147                         \\
$\;\;\;\;\;\;\;\;\;$ AUX                                                                        & 0.25                                                                                     & 2069                         \\
$\;\;\;\;\;\;\;\;\;$ ADP                                                                        & 0.23                                                                                     & 2476                         \\
$\;\;\;\;\;\;\;\;\;$ DET                                                                        & 0.20                                                                                      & 3265                         \\
\multicolumn{2}{l}{Q3b. Related words - per POS group} \\
$\;\;\;\;\;\;\;\;\;$ Noun Group                                                          & 0.60                                                                                      & 5478                         \\
$\;\;\;\;\;\;\;\;\;$ Verb Group                                                                 & 0.32                                                                                     & 4999                         \\
$\;\;\;\;\;\;\;\;\;$ ADJ Group     & 0.29      & 4584\\[3pt] 
\multicolumn{2}{l}{Q4. Related Subjects and Objects}    &\\ 
$\;\;\;\;\;\;\;\;\;$ Subject       & 0.29      & 1611\\ 
$\;\;\;\;\;\;\;\;\;$ Object        & 0.43      & 1618\\ 
\hline
\end{tabular}
\caption{\label{tab:pos_results} Correlation between 
features and the relatedness of sentence pairs. A rule of thumb for interpreting the numbers:  
0--0.19: very weak; 0.2--.39: weak; 0.4--0.59: moderate; 0.6--0.79: strong; 0.8--1: very strong.}
\vspace*{-3mm}
\end{table}

\subsection{Results}
\begin{figure}[tbp]
 \centering
 \includegraphics[width=0.45\textwidth]{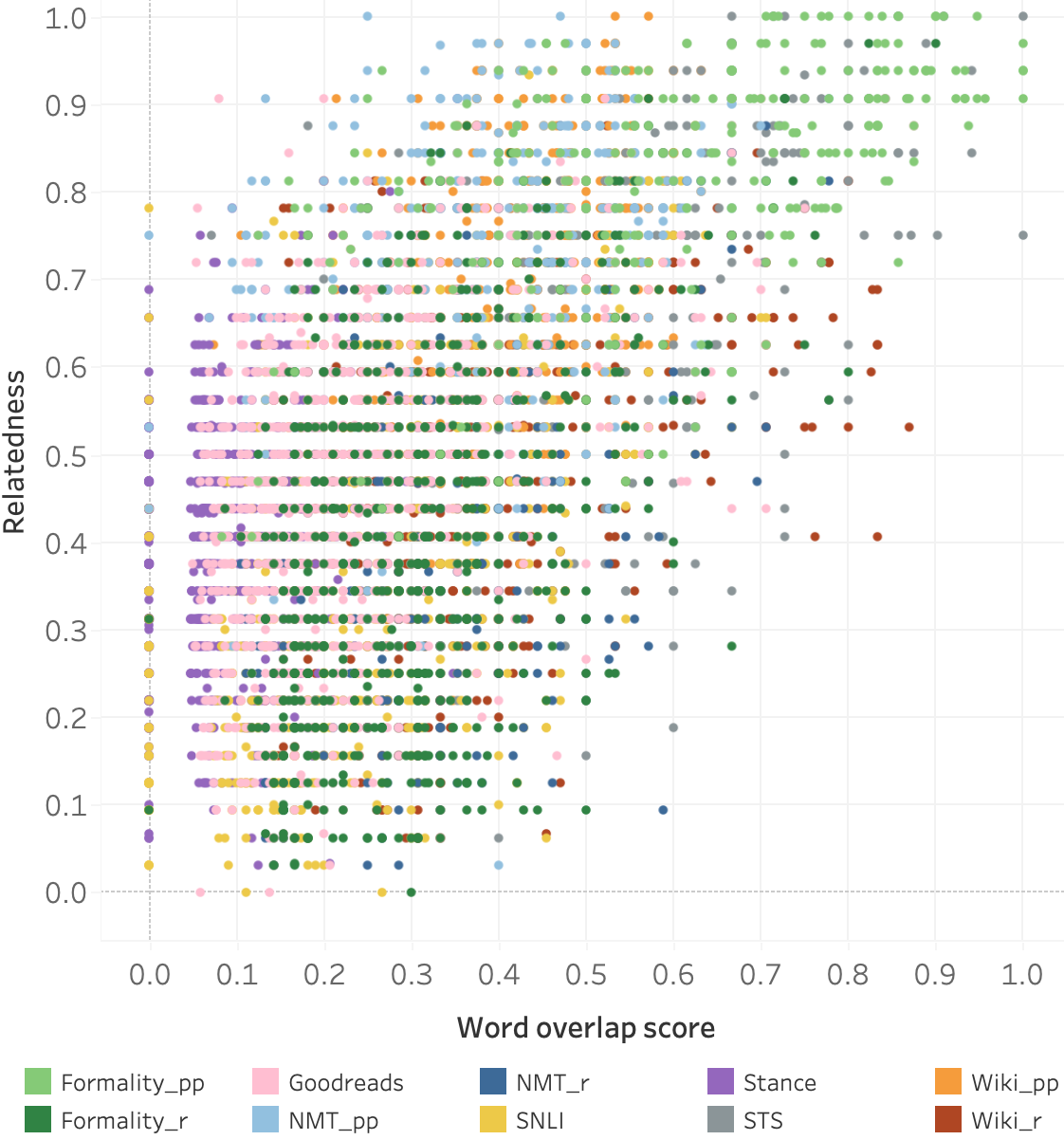}
 \caption{\label{img:str_woverlap} Scatter plot showing the relationship between lexical overlap and semantic relatedness of sentence pairs. Each dot in the plot is a sentence pair.}
 \vspace*{-3mm}
 \end{figure}

Table \ref{tab:pos_results} shows the results. Row Q1 shows that simple word overlap obtains a correlation of 0.57 (considered to be at the 
\gr{high} end of 
\gr{weak} correlation). Figure \ref{img:str_woverlap} is a scatter plot where the x-axis is the word overlap score, the y-axis is the relatedness score, and each dot is a sentence pair. Observe that a number of pairs fall along the diagonal; however, there are also a large number of pairs along the top-left side of this diagonal. This suggests that even though STR-2022 has pairs where the relatedness increases linearly with the amount of word overlap, there are also a number of pairs where a small amount of word overlap results in substantial amount of relatedness. The 
sparse bottom-right side of the plot indicates that it is rare for there to be substantial word overlap, and yet very low relatedness.
On average,  occurrence of related words across a sentence pair leads to 
\gr{slightly higher} 
relatedness scores 
\gr{than} lexical overlap (row Q2). 

The Q3a rows in Table \ref{tab:pos_results} show correlations for related tokens of a given part of speech.\footnote{Only those POS tags that occur in both sentences of a pair in more than 10\% of the pairs  are considered ($>$550 pairs).} (The rows are in order from highest to lowest correlation.) Observe that proper nouns (PROPN) and nouns have the highest numbers. It is somewhat surprising that related verbs do not contribute greatly to semantic relatedness; they have similar correlations as pronouns and adverbs, and markedly lower than adjectives and nouns. 
Not surprisingly, 
determiners (DET) 
are at the lower end of  
weak correlation. 

The Q3b rows show correlations of coarse POS categories: NOUN Group (NOUN, PRON, PROPN), VERB Group (VERB, AUX), and ADJ Group (ADJ, ADP, ADV). 
We see that presence of related nouns in a sentence pair impacts semantic relatedness much more than any other POS group.


Since related nouns were found to be especially important, we also wanted to determine what impacts overall relatedness more: the presence of related nouns in the subject position  or in the  object position. Q4 rows show that, on average, related objects lead to markedly higher sentence-pair relatedness than related subjects. 

In order to examine whether lexical overlap and some 
POS are less or more relevant in low or high relatedness pairs, we repeated the experiment of Table 4, only for pairs with relatedness scores $<$0.5, and separately, only for pairs with scores $\geq$0.5. We find that for the $<$0.5 relatedness pairs, only the existence of related proper nouns across sentence pairs has moderate correlation with the semantic relatedness of sentences; the correlation is weak for nouns, and close to 0 for all other parts of speech.  
The notable importance of related proper nouns and nouns is likely because they indicate a common topic, person,  or object being talked about in both sentences---making the two sentence pairs related. 
For the $\geq$0.5 relatedness pairs, the correlations are weak for most POS; highest for nouns; and the gap between nouns and adjectives, adverbs, and verbs is reduced. Lexical overlap in general has a much higher correlation for the $\geq$0.5 relatedness pairs than the $<$0.5  pairs. Detailed results are in Appendix \ref{ref:appendix_features}.     \\[-10pt]


\section{Evaluating Sentence Representation Models using STR-2022}
\label{sec:sentenceRep}
Since STR-2022 captures a wide range of fine-grained relations that exist between sentences, it is a valuable asset in evaluating sentence representation and embedding models. Essentially, predicting semantic relatedness is treated as a regression task, where first, using various unsupervised and supervised approaches described in the two sub-sections below, we represent each sentence as a vector. We use the cosine similarity between the vectors as a prediction of their semantic relatedness. We use the Spearman correlation between the prediction and gold relatedness scores to measure the goodness of the relatedness predictions (and in turn of the sentence representation).

The experiments below (unless otherwise specified) all involve 5-fold cross-validation (CV) on STR-2022. We report the average of the Spearman correlations across the folds. Note that even for models that do not require training (e.g., Dice score), to enable direct comparisons with trained methods, we evaluate their performance on each test fold independently and report the average of the correlations across folds.

\begin{table}[tbp]
\centering
\small
\begin{tabular}{lr}
\hline
\multicolumn{1}{l}{\textbf{Model}}   & \textbf{\begin{tabular}[c]{@{}c@{}}Spearman\end{tabular}} \\ \hline
\textit{Baseline} &\\
$\;\;\;\;$ 1. Lexical overlap (Dice) & 0.57                         \\ 
\textit{Unsupervised, Static Embeddings} &\\
$\;\;\;\;$ 2. Word2Vec (mean, Googlenews)          & 0.60                         \\ 
$\;\;\;\;$ 3. Word2Vec (max, Googlenews)          & 0.54                         \\ 
$\;\;\;\;$ 4. GloVe (mean, Common Crawl)           & 0.49                         \\ 
$\;\;\;\;$ 5. GloVe (max, Common Crawl)           & 0.56                         \\ 
$\;\;\;\;$ 6. GloVe (mean, 200\_Twitter)           & 0.44                         \\ 
$\;\;\;\;$ 7. GloVe (max, 200\_Twitter)           & 0.48                       \\ 
$\;\;\;\;$ 8. Fasttext (mean, Common crawl)        & 0.29                    \\ 
$\;\;\;\;$ 9. Fasttext (max, Common crawl)        & 0.24                         \\
\textit{Unsupervised, Contextual Embeddings}&\\
$\;\;\;\;$ 10. BERT-base (mean)                        & 0.58                        \\ 
$\;\;\;\;$ 11. BERT-base (max)                          & 0.55                          \\ 
$\;\;\;\;$ 12. BERT-base (cls)                          & 0.41                        \\ 
$\;\;\;\;$ 13. RoBERTa-base (mean)                      & 0.48                         \\ 
$\;\;\;\;$ 14. RoBERTa-base (max)                       & 0.47                        \\ 
$\;\;\;\;$ 15. RoBERTa-base (cls)                       & 0.41                        \\ 
\multicolumn{2}{l}{\textit{Supervised (Fine-tuning on portions of STR-2022)}} \\
$\;\;\;\;$ 16. BERT-base (mean)        & 0.82 \\
$\;\;\;\;$ 17. RoBERTa-base (mean)        & 0.83 \\ \hline
\end{tabular}
\caption{\label{tab:unsupervised_embeddings} Average correlation between human annotated relatedness of sentence pairs and the cosine distance between their embeddings across the CV runs.}
\vspace*{-3mm}
\end{table}

\subsection{Do Unsupervised Embeddings Capture Semantic Relatedness?}
\label{sec:unsupervised_embed}
We first explore unsupervised approaches to sentence representation where the embedding of a sentence is derived from that of its constituent tokens. The token embedding can be of two types:
\vspace*{-2mm}
\begin{itemize}
    \item \textbf{Static Word Embeddings:} 
    We tested three popular models: Word2Vec \cite{mikolov2013linguistic}, GloVe \cite{pennington2014glove}, and Fasttext \cite{grave2018learning}.
\vspace*{-2mm}
    \item \textbf{Contextual Word Embeddings:} We tested pretrained contextual embeddings from BERT \cite{devlin2019bert} and RoBERTa \cite{liu2019roberta}. We use the bert-base-uncased and roberta-base models from the HuggingFace library.\footnote{\url{https://huggingface.co}}  
    \vspace*{-2mm}
\end{itemize}
\noindent We obtain sentence embeddings by both mean-pooling and max-pooling the token embeddings from the final layer. For the contextual embeddings, we also explore using the embedding of the classification token ([CLS]).

Table \ref{tab:unsupervised_embeddings} shows the results. As baseline, we include how well simple lexical overlap (Dice score) predicts relatedness (row 1). Observe that mean-pooling with word2vec (row 2) obtains slightly higher correlation than the baseline, but the majority of the static embedding models fail to obtain better correlations (rows 3--9). The contextual embeddings from BERT and RoBERTa do not perform better than the word2vec embeddings (rows 10--15). Overall, the unsupervised 
methods leave much room for improvement.

\subsection{Do Supervised Embeddings Capture Semantic Relatedness?}
\label{sec:supervised_embed}

We now evaluate the performance of BERT-based models on STR-2022 when formulated as a \textit{supervised} regression task. We use the S-BERT cross-encoder framework of \citet{reimers-2019-sentence-bert}, and apply mean-pooling on top of the token embeddings of the final layer to obtain sentence embeddings. The model is trained using a cosine-similarity loss---the cosine between the embeddings of a sentence pair is compared to the gold semantic relatedness scores to obtain the Mean Squared Error (MSE) loss for each datapoint.

Table \ref{tab:unsupervised_embeddings} rows 16 and 17 show the results: fine-tuning on STR-2022 leads to considerably better relatedness scores.

\begin{table}[tbp]
\centering
\small
\begin{tabular}{lccc}
\hline
    &\textbf{Dice}  & \multicolumn{2}{c}{\bf SBERT(RoBERTa)}\\
                 &CV & CV & LOO CV \\ \hline \\[-9pt]
    STS                & 0.60  & 0.79 & 0.82\\ 
    SNLI               & 0.53 & 0.80 & 0.77\\ 
    Stance             & 0.20 & 0.49 & 0.39\\ 
    Goodreads          & 0.44 & 0.73 & 0.70\\ 
    Wiki               & 0.48 & 0.79 & 0.75\\ 
    Formality           & 0.69 & 0.86 & 0.83\\ 
    ParaNMT             & 0.44 & 0.80 & 0.79\\  \hline
\end{tabular}
\caption{\label{tab:supervised_embeddings} Breakdown of 
average test-fold 
correlations for each source: (a) using lexical overlap (Dice),
(b) using SBERT and some in-domain data for fine-tuning (in addition to data from other domains), and 
(c) using SBERT and only out-of-domain data for fine-tuning (LOO CV).
CV: cross-validation. LOO: leave-one-out.}
\vspace*{-3mm}
\end{table}

\subsubsection{Impact of Domain on Fine-Tuning}
The results above show that fine-tuning is critical for better sentence representation. However, it is well-documented that the domain of the data can have substantial impact on results; especially when quite different from the training data. With the inclusion of data from various domains in STR-2022 (Table \ref{tab:data_sources}), one can systematically explore performance on individual domains, as well as the extent to which performance may drop if no training data from the target domain is included for training.

Table \ref{tab:supervised_embeddings} shows the results. The RoBERTa CV column shows a breakdown of results 
by source (domain). Essentially, these are results for the scenario where some portion of in-domain data is included in the training folds (along with data from other domains), and the system correlations are determined only on the test fold's target domain pairs. Observe that performance on most domains is comparable to each other. 

The LOO CV column shows correlations with a leave-one-out cross-validation setup: no in-domain training data is used and system correlations are determined only for the target domain pairs. Observe that this leads to drops in scores for all domains except STS. 
However, the drop is small; 
and scores are still much higher than the lexical overlap (Dice CV) baseline. This suggests that the diversity of data in the remaining subsets is useful in overcoming a lack of in-domain training data.

\section{Utility of Semantic Relatedness and STR-2022 in Downstream NLP Tasks}
\label{sec:utility}
\bl{Semantic relatedness is central to textual coherence and narrative structure. Often, sentences in a document 
are not paraphrases, entailments, or similar,  but rather semantically related to each other. This need for continuity of meaning has long been identified as a crucial component of language \cite{halliday1976cohesion,morris1991lexical}. 
Thus, when generating a summary or a response to a question, systems must choose sentences that are \textit{not} paraphrases or entailments of each other, but yet suitably semantically related. Therefore, being able to judge both similarity and relatedness is crucial. 

Since we made STR-2022 publicly available, it has already been used in some projects. Notable among these is \citet{wang2022just}.
\citet{wang2022just} propose a new intrinsic evaluation method, \textit{EvalRank}, that focuses on local neighborhoods 
(how well systems identify close neighbors, rather than how well they rank the full set of pairs). 
Using STR-2022, they are able to obtain markedly higher correlations between performance scores on the intrinsic evaluation
and performance on downstream tasks (seven NLP tasks including NLI, question classification, caption retrieval, and sentiment analysis).
Their ablation study demonstrates that using STS instead of STR-2022 decreases performance up to 10 points, leading them to conclude that  
STR-2022 is particularly useful in generating sentence embeddings for downstream tasks.}

\section{Conclusion}

We created STR-2022, the first dataset of English sentence pairs annotated with fine-grained relatedness scores. We used a comparative annotation method that produced a split-half reliability of 0.84. Thus, we showed that speakers of a language can reliably judge semantic relatedness. We used the dataset to explore several research questions pertaining to what makes two sentences more related. 
Finally, we used STR-2022 to evaluate the ability of sentence representation methods to embed sentences in vector spaces such that those that are closer to each other in meaning are also closer in the vector space. The dataset is made freely available; facilitating  further research in semantic relatedness and sentence representation.

\section{\fl{Limitations}}
\label{sec:limitations}

\fl{For limitations surrounding the dataset, please refer to the Ethics Statement (\S \ref{sec:ethics}).
} 
\fl{ 
In our experiments, we used the most common methods for sentence representations (e.g., mean-pooling and max-pooling of traditional and contextual word embeddings). However, there may exist other embeddings which are better suited for predicting semantic relatedness (e.g., other order-aware embeddings). Expanding the set of embedding techniques tested using our dataset may yield different results and provide us a stronger understanding of the effects of different representation techniques.}
\fl{Furthermore, 
while we explored the impact of some sentence-pair features such as lexical overlap, POS, and 
some aspects of sentence structure
(subject and object) on semantic relatedness, 
we did not explore the impacts of other features such as logicality and common sense reasoning on relatedness. These remain interesting directions for future work.}

\section{Ethics Statement}
\label{sec:ethics}
This paper respects existing intellectual property by making use of only publicly and freely available datasets.  The crowd-sourced task was approved by our Institutional Research Ethics Board. The annotators were based in the United States of America and were paid the federal minimum wage of \$7.25 per hour. 
Our annotation process stored no information about annotator identity and as such there is no privacy risk to them.  The individual sentences selected did not have any risks to privacy either (as evaluated by manual annotation of the sentences). Models trained on this dataset may not generalize to external datasets gathered from different populations. Knowledge about language features may not generalize to other languages.

Any dataset of semantic relatedness entails several ethical considerations. 
We list some notable ones below. \fl{Many of these were first introduced in the context of sentiment lexicons \cite{mohammad2020practical,mohammad2023LexEthics}. We adapted them to semantic relatedness datasets and added to the discussion.}
\vspace*{-1mm}
\begin{itemize}
    \item \textit{Coverage:} We sampled English sentences from a diverse array of sources from the internet, with a focus on social media. Yet, it is likely that several types of sentences (and several demographic groups) are not well-represented in STR-2022. The dataset likely includes more sentences by people from the United States and Europe and with a socio-economic and educational backgrounds that allow for social media access.
    \vspace*{-1mm}
    \item \textit{Not Immutable:} The relatedness scores do not indicate an inherent unchangeable attribute. The relatedness can change with time, but the dataset entries are largely fixed. They pertain to the time they are created.
    \vspace*{-1mm}
    \item \textit{Socio-Cultural Biases:} The annotations of relatedness capture various human biases. These biases may be systematically different for different socio-cultural groups. Our data was annotated by US annotators, but even within the US there are diverse socio-cultural groups.
    \vspace*{-1mm}
    \item \textit{Inappropriate Biases:} Our biases impact how we view the world, and some of the biases of an individual may be inappropriate. For example, one may have race or gender-related biases that may percolate subtly into one's notions of how related two units of text are. 
    Our dataset curation was careful to avoid sentences from problematic sources, and we have not seen any inappropriate relatedness judgments, but it is possible that some subtle inappropriate biases still remain.
    Thus, as with any approach for sentence representation or semantic relatedness, we caution users to explicitly check for such biases in their system regardless of whether they use STR-2022.  
    \vspace*{-1mm}
    \item \textit{Perceptions (not “right” or “correct” labels):} Our goal here was to identify common perceptions of semantic relatedness. These are not meant to be ``correct'' or ``right'' answers, but rather what the majority of the annotators believe based on their intuitions of the English language.
    \vspace*{-5mm}
    \item \textit{Relative (not Absolute):} The absolute values of the relatedness scores themselves have no meaning. The scores help order sentence pairs relative to each other. For example, a pair with a higher relatedness score should be considered more related than a pair with a lower score. No claim is made that the mid-point (relatedness score of 0.5) separates related words from unrelated words. One may determine categories such as \textit{related} or \textit{unrelated} by finding thresholds of relatedness scores optimal for their use/task.
\end{itemize}
\noindent We recommend careful reflection of ethical considerations relevant for the specific context of deployment when using STR-2022.



\bibliography{custom}
\bibliographystyle{acl_natbib}

\newpage
\appendix

\section{\gr{Further Details on Sampling Sentence Pairs from Source Datasets}}
\label{ref:appendix_data}
This 
Appendix provides further information about the sources of data and how sentence \gr{pairs} were sampled from them to be included in STR-2022.

\subsection{STS Data}
We selected 250 sentence pairs from existing STS corpora. This selection was done to enable a small investigation into the interplay between relatedness and similarity, which could serve as motivation for further investigation in future work. For this dataset, we randomly sampled 50 sentence pairs from each of bucket of annotation (i.e., 50 sentence pairs having an STS similarity scores falling in $[0,1)$, 50 sentence pairs having scores in $[1,2)$, and so on).

\subsection{Stance Data}
We created 750 sentence pairs by sampling from \citet{mohammad2016dataset}'s dataset of tweets labeled for stance. The original dataset is composed of individual tweets labelled for both stance (`For', `Against', `Neither Inference Likely') and sentiment (`Positive', `Negative', `Neutral'). The dataset was built from tweets focused on six targets: `Atheism', `Climate Change', `Donald Trump', `Feminism', `Hillary Clinton', `Abortion'.  

When curating our sentence pairs, we limited the possible targets to `Hillary Clinton', `Donald Trump', and `Abortion'. Sentence pairs were chosen such that both sentences shared the same target. 500 sentence pairs shared their stance towards their target (\gr{ i.e., 250 \textit{for--for} pairs and 250 \textit{against--against} pairs}). 250 sentences pairs differed on their stance \gr{(i.e., \textit{250 for--against} pairs)}. We did not use any lexical overlap heuristic to specify which tweets should be paired with each other because we were interested in studying whether overlap in topic was a strong enough signal to impact relatedness. That is, by choosing pairs with the same target, we were already pre-selecting for various degrees of relatedness.

\subsection{SNLI Data}
\label{snli-data-desc}
We created  750 sentence pairs by sampling from the Stanford Natural Language Inference (SNLI) Dataset \cite{snli}. SNLI is composed of image description captions; for each caption, multiple premise sentences are generated, along with multiple possible hypothesis sentences that could possibly belong to each premise. To build our sentence pairs we sought to pair different premise sentences together. We did not wish to pair between premise and hypothesis sentences as the sentence structure was significantly different (and simpler for the hypothesis sentences), as noted by the creators of the dataset. Even still, the majority of premise sentences are very short (with a mean token count of 14), often following very simple (and similar) grammatical structure. 

To generate the sentence pairs, first we removed all sentences with less than 5 or more than 25 tokens. Then, for each token in all remaining sentences, we replaced each token with its most frequent synonym, using Roget's Thesaurus \cite{roget1911roget} to define synonymous relationships. Words which did not have synonyms were left unchanged.    The intention behind replacing each word with its most frequent synonym was to ensure that synonymous phrasings would count as overlaps when we measure it. We then randomly selected 750 sentences to serve as the first sentence of our final pairings. To find the second sentence to each pairing we looped through all premise sentences and returned the first sentence that satisfied two conditions: 1) The unigram overlap was greater than or equal to 25\% and less than 75\% of the first sentence, and 2) the difference in length between both sentences did not exceed 25\%.

\subsection{Wikipedia Data}
\label{wiki-data-desc}
We sampled 1000 sentence pairs from a dataset that pairs sentences from English Wikipedia with sentences from Simple English Wikipedia. Created to enable the task of sentence simplification, the paired sentences, paired using rules-based classification, are often very closely related. We used this dataset in two ways: 1. Extracting sentence pairs which serve as paraphrases or near paraphrases (we refer to these as Wiki\_pp), and 2. pairing sentences to other random sentences in the dataset (we refer to these as Wiki\_r).\\[-10pt]

\noindent \textbf{Wiki\_pp}:
First, we removed any pairings for which either sentence was less than 5 words or more than 25 words. Then we narrowed the list of pairings further by removing any pairings that did not share more than 25\% but less than 75\% of unique unigrams. From the remaining sentence pairs, we randomly selected 500 paired sentences. \\[-10pt]

\noindent \textbf{Wiki\_r}:
Here, we only made use of the full sentences from the original Wikipedia, discarding sentences from Simple Wikipedia. We removed all sentences that have less than 5 or more than 25 tokens. To create the sentence pairs, we looped in a random order through all possible pairing of sentences. We paired two sentences if they share at least 25\% of their tokens but less than 75\% of their tokens AND the difference in length between both sentences did not exceed 25\%. We stopped once we had generated 500 sentence pairs.

\subsection{Goodreads Data}
\label{gd-data-desc}
We created 1000 sentence pairs by sampling from the UCSD Goodreads Dataset \citep{DBLP:conf/recsys/WanM18, DBLP:conf/acl/WanMNM19}, which has book reviews from the Goodreads website. We limited the sampling to the `Fantasy and Paranormal' genre, since it contained a relatively higher number of reviews per book, allowing for a higher possibility of sampling more related sentence pairs. Each review was first split into sentences using the default NLTK sentence tokenizer; we kept only those sentences with the number of tokens between 5 and 25. We then randomly examined pairs of sentences, and quantified the lexical overlap between then with an IDF-weighted Dice overlap score. The pairs were then assigned to buckets based on this overlap score; the range of each bucket was obtained by first finding 50 equally-spaced percentiles of the entire score distribution. We then sampled exponentially increasing number of sentences from low to high weighted Dice overlap bins such that a total of 1000 sentence pairs were included.

\subsection{ParaNMT Data}
ParaNMT \citep{wieting-gimpel-2018-paranmt} is a dataset of 51 million sentential paraphrases that were automatically generated using a neural machine translation system. We generated two sets of pairs from these sentences corresponding to paraphrases and random pairs:\\[-10pt]

\noindent \textbf{ParaNMT\_pp:}
We assigned paraphrases to buckets based on the Dice score  between the two sentences. We divided the range of scores into 100 equally-sized percentiles. We then sampled pairs uniformly from each bucket, for a total of 450 sentence pairs.\\[-10pt]

\noindent  \textbf{ParaNMT\_r:}
For the random, non-paraphrase sentence pairings, we used the Dice score to extract 300 pairs, analogous to the creation of the \textbf{Wiki\_r} pairs.

\subsection{Formality Data}
Our third paraphrase corpus is the Formality dataset from \citet{rao2018dear} (They refer to it as GYAFC). This consists of human-written formal and informal paraphrases for sentences sourced from the Yahoo! Answers platform. Our sampling procedure for this dataset followed that of the ParaNMT dataset.\\[-10pt]

\noindent \textbf{Formality\_pp:}
We assigned sentences to one of 50 buckets based on their lexical overlap score as before. We then uniformly sampled from each bucket to extract 300 sentence pairs.\\[-10pt]

\noindent \textbf{Formality\_r:}
We sampled random pairings of sentences using the token overlap and length difference conditions as defined for Wiki\_r and ParaNMT\_r. We extracted 700 such sentence pairs.

\begin{table*}[t]
\centering
\small
\begin{tabular}{lrrr}
\hline
                            &\multicolumn{3}{c}{\textbf{Spearman}} \\ 
 \textbf{Question}          &0--1 pairs & $<$0.5 pairs     &$\geq$0.5 pairs\\ \hline\\[-9pt]

Q1. Lexical overlap       &    0.57       & 0.14    & 0.52\\ [3pt]
Q2. Related words - All       & 0.61          & 0.14  &0.50\\ [3pt]
\multicolumn{2}{l}{Q3a. Related words - per POS} \\
$\;\;\;\;\;\;\;\;\;$ PROPN          & 0.50  &0.34   &0.26          \\
$\;\;\;\;\;\;\;\;\;$ NOUN           & 0.45  &0.18   &0.37                        \\
$\;\;\;\;\;\;\;\;\;$ ADJ            & 0.36  &0.04   &0.35       \\
$\;\;\;\;\;\;\;\;\;$ VERB           & 0.31  &0.03   &0.31               \\
$\;\;\;\;\;\;\;\;\;$ PRON           & 0.30  & 0.01  & 0.30\\
$\;\;\;\;\;\;\;\;\;$ ADV            & 0.28  & 0.04  & 0.35  \\
$\;\;\;\;\;\;\;\;\;$ AUX            & 0.25  & 0.03  & 0.20\\
$\;\;\;\;\;\;\;\;\;$ ADP            & 0.23  & 0.07    & 0.22\\
$\;\;\;\;\;\;\;\;\;$ DET            & 0.20  & 0.03  & 0.19     \\ [3pt]
\multicolumn{2}{l}{Q3b. Related words - per POS group} \\
 $\;\;\;\;\;\;\;\;\;$ Noun Group        & 0.60  & 0.34   & 0.41\\
 $\;\;\;\;\;\;\;\;\;$ Verb Group        & 0.32  & 0.09   & 0.29\\
 $\;\;\;\;\;\;\;\;\;$ ADJ Group         & 0.29  & 0.04   & 0.32\\[3pt] 
\multicolumn{2}{l}{Q4. Related Subjects and Objects}    &\\ 
 $\;\;\;\;\;\;\;\;\;$ Subject       & 0.29      &0.00 & 0.32\\ 
 $\;\;\;\;\;\;\;\;\;$ Object        & 0.43      &0.14 & 0.33\\ 
\hline
\end{tabular}
\caption{\label{tab:pos_results_split} Correlation between 
features and the relatedness of sentence pairs in STR-2022 when considering full relatedness range (0--1), only the pairs with relatedness $<0.5$, and only the pairs with relatedness $\geq0.5$.\\
Note: The 0--1 pairs column was shown earlier in Table 4. It is repeated here for ease of comparison.}
\end{table*}

\section{Correlation of Features in Low and High Relatedness Sentence Pairs}
\label{ref:appendix_features}
As discussed in Section 5.2, in order to examine whether lexical overlap and some parts of speech are less or more relevant in low or high relatedness pairs, we repeated the experiment in Table 4, only for pairs with relatedness scores less than 0.5 and also for pairs with scores greater than 0.5. 
Table \ref{tab:pos_results_split} shows the detailed correlation scores. See Section 5.2 for a discussion of the main trends.

\section{\fl{Sample Sentence Pairs from STR-2022}}
Table \ref{tab:sample-pairs} presents sample sentene pairs from different domains.

\begin{table*}[h]
    \centering
    \small
    \begin{tabular}{llr}
    \textbf{Source} & \textbf{Sentence Pairs} & \textbf{STR score} \\ \hline 
    Formality\_pp & \makecell{\textit{I think Taylor is really cute, but I hate his voice.} \\
\textit{I think Taylor is SUPER cute...but I hate his voice.}} & 1.000 \\
\hline
    Wiki\_pp & \makecell{\textit{It is sometimes referred to as the trunk.} \\
    \textit{Some people also call it the trunk.}} & 0.969 \\
    \hline
    Goodreads & \makecell{\textit{I loved this short story - wish it were longer!} \\ 
    \textit{It was a quick read and part of me wished that it would go on a little longer.}} & 0.844 \\
    \hline
    Wiki\_r & \makecell{\textit{On August 2 , a tropical storm hit Northeastern Florida .} \\
    \textit{In early October , a hurricane caused damage and erosion to northeastern Florida .}} & 0.625 \\
    \hline
    Stance & \makecell{\textit{So unfortunate \#thebriefcase @cbs. Adoption isn't always the answer.} \\ 
    \textit{Just remember, there is a living family out there just waiting to \#adopt your aborted baby.}} & 0.562 \\ 
    \hline
    SNLI & \makecell{\textit{A woman in speaking in a theater.} \\
    \textit{deleon speaking into a mic.}} & 0.406 \\
    \hline
    ParaNMT\_pp & \makecell{\textit{Are you--are you going to tell every one?} \\ 
    \textit{will you say it now -- all of you?}} & 0.334 \\ 
    \hline
    Formality\_r & \makecell{\textit{i believe in american dreams ...} \\
    \textit{You are the woman of my dreams}} & 0.219 \\
    \hline
    STS & \makecell{\textit{A person is riding a horse.} \\ 
    \textit{A woman is slicing potatoes.}} & 0.062 \\
    \hline
    \end{tabular}
    \caption{\label{tab:sample-pairs} Sample sentence pairs from different domains in the STR-2022 dataset.}
\end{table*}
\end{document}